\documentclass[lettersize,journal]{IEEEtran}
\usepackage{amsmath,amsfonts}
\usepackage{algorithmic}
\usepackage{algorithm}
\usepackage{array}
\usepackage[caption=false,font=normalsize,labelfont=sf,textfont=sf]{subfig}
\usepackage{textcomp}
\usepackage{stfloats}
\usepackage{url}
\usepackage{verbatim}
\usepackage{graphicx}
\usepackage{cite}
\usepackage{multirow}
\usepackage{multicol}
\usepackage{float}
\usepackage{color}
\usepackage{hyperref}

\usepackage{orcidlink}
\usepackage{tikz}
\hypersetup{
colorlinks=true,
linkcolor=black,
citecolor=black
}

\begin{document}

\title{YUNet: Improved YOLOv11 Network for Skyline Detection}

\author{Gang Yang\href{https://orcid.org/0000-0001-6096-9401}{\includegraphics[scale=0.08]{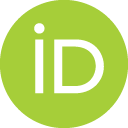}}, {Miao Wang}, {Quan Zhou\href{https://orcid.org/0009-0001-1240-9415}{\includegraphics[scale=0.08]{ORCIDiD.png}}}, {Jiangchuan Li\href{https://orcid.org/0009-0005-3904-1007}{\includegraphics[scale=0.08]{ORCIDiD.png}}}
\thanks{This paper is supported by Ministry of Natural Resources-Provincial Cooperation Project: Research on the Management Platform and Application of Overall Protection and Revitalization of Cultural and Natural Heritage (No.: 2024ZRBSHZ055)}
\thanks{Manuscript received April 19, 2021; revised August 16, 2021.}}

\markboth{Journal of \LaTeX\ Class Files,~Vol.~14, No.~8, August~2021}%
{Shell \MakeLowercase{\textit{et al.}}: A Sample Article Using IEEEtran.cls for IEEE Journals}

\IEEEpubid{0000--0000/00\$00.00~\copyright~2021 IEEE}

\maketitle

\begin{abstract}
Skyline detection plays an important role in geolocalizaion, flight control, visual navigation, port security, etc. The appearance of the sky and non-sky areas are variable, because of different weather or illumination environment, which brings challenges to skyline detection. In this research, we proposed the YUNet algorithm, which improved the YOLOv11 architecture to segment the sky region and extract the skyline in complicated and variable circumstances. To improve the ability of multi-scale and large range contextual feature fusion, the YOLOv11 architecture is extended as an UNet-like architecture, consisting of an encoder, neck and decoder submodule. The encoder extracts the multi-scale features from the given images. The neck makes fusion of these multi-scale features. The decoder applies the fused features to complete the prediction rebuilding. To validate the proposed approach, the YUNet was tested on Skyfinder and CH1 datasets for segmentation and skyline detection respectively. Our test shows that the IoU of YUnet segmentation can reach 0.9858, and the average error of YUnet skyline detection is just 1.36 pixels. The implementation is published at \url{https://github.com/kuazhangxiaoai/SkylineDet-YOLOv11Seg.git}.
\end{abstract}


\section{Introduction}
\IEEEPARstart{T}{he} skyline is defined as the boundary between sky and non-sky region.  The skyline detection plays a great role in geolocalizaion, flight control, visual navigation, port security, forest fire prevention, etc.
Therefore, the development of skyline detection technology has draw more and more attention in recent years.
\par Although the skyline detection is significant valuable in many fields, its practical performance is limited by the complicated environment, such as variety of weather and illumination. As results of weather and illumination changes, the images captured at the same place and view angle may be different obviously. Therefore, it is necessary to develop the skyline detection algorithms working in complicated environment. 
\par Accordingly, we proposed the YUNet algorithm for skyline detection in the complicated environments, which developed from YOLOv11 and UNet architecture. The YOLOv11 architecture is a state-of-art approach for object detection, but which cannot provide the ability of semantic segmentation. The UNet is designed for the semantic segmentation task with convolutional networks. It is limited by the receptive field of these convolutional networks, causing the lack of large range contextual information. To overcome the shortage of YOLOv11 and UNet architecture, the YUNet that improved the YOLOv11 to an UNet-like architecture is proposed. The YUNet consists of encoder, neck and decoder submodule. Similarly to the Unet architecture, the encoder and decoder is used for feature extraction and prediction rebuilding, respectively. The neck developed from the PAFPN of YOLOv11 architecture, which makes fusion of the multi-scale features and captures the large range contextual information.
\par The YUNet is a image segmentation model. It can decide each pixel in the given image whether belongs to sky region or not. Based on the image segmentation, we adopt the Canny algorithm to extract the segmentation boundary as the skyline.
\par For evaluation of the YUNet performance on both sky segmentation and skyline detection tasks, we test it on two public datasets including Skyfinder and CH1 dataset. The Skyfinder dataset is adopt for validation of the YUNet on the sky segmentation task. The IoU of YUNet can reach 0.9858 for test set of Skyline dataset, when it is performing the sky segmentation task. The CH1 dataset is adopt for evaluation about skyline detection task. The average error of YUNet is just 1.36 pixels, when it is performing the skyline detection task.

\section{Related work}
\IEEEpubidadjcol
The skyline detection algorithms can be grouped by two main categories. The first group methods are mainly designed based on feature descriptor and machine learning technologies. These methods are proposed for a long time, so they are regarded as traditional methods. For example, Cornall \textit{et al.}\cite{iet:/content/journals/10.1049/el_20060547}, based on the analysis of the sky RGB channels, employed blue as distinctive feature to extract the boundary of sky and non-sky region. Chiu \textit{et al.} \cite{Chiu} extracted the skyline curve under the brightness and contrasts information. Lie \textit{et al.} \cite{LIE2005221} obtained the edge map from the given image. Subsequently, they constructed the multi-stage map from the edge map for skyline detection. Dong \textit{et al.} \cite{Dong2010} calculated the local maximum grayscale complexity and multiplied it by a parameter which is less than one to obtain a threshold for image binarization. They identified regions with relatively high complexity, such as the sea-to-sky convergence line, and used HT to find the skyline.. Fefilatyev \textit{et al.} \cite{Fefilatyev2006} took the texture and color feature as the input of support vector machine (SVM) to judge pixels whether belong to sky region or not. Ahmad \textit{et al.} \cite{Ahmad2015} proposed a skyline detection algorithm using machine learning and Dynamic Programming (DP) to extract the skyline from the classification map. In this approach, each pixel is assigned a classification score as a likelihood of the pixel belonging to the horizon line, and representing the classification map as a multi-stage graph. Using DP, the horizon line can be extracted by finding the path that maximizes the sum of classification scores. Ahmad \textit{et al.} \cite{Ahmad2021} proposed a bunch of learned filters based on the local structure tensors, and applied these filters to the patches around them for generation of skyline predict score. Generally speaking, the traditional methods require manual design of image feature, and a few labeled data for the training of classifiers or filters. Because of  unreliability of feature design and small amount of model parameters, the robustness of traditional methods are usually unsatisfactory for variety of lightness and weather. 
\par The recent second group methods are mainly based on deep learning technologies, which employs trainable deep neural networks for skyline extraction. For example, Verbickas and Whitehead \cite{Verbickas2014} applied convolutional neural networks (CNNs) to the detection of the horizon, training them to detect the sky and ground regions and the horizon line in flight videos. Poriz \textit{et al.} \cite{Porzi2016} proposed a sky segmentation method based on improved U-Net architecture\cite{Ronneberger2015}, which introduced intermediate levels of supervision to support the learning process. Frajberg \textit{et al.} \cite{Frajberg2017}  presented the results of training a CNN for extracting mountain skylines. Yang \textit{et al.} \cite{Yang2021} detected the sea-sky-line based on the improved YOLOv5 algorithm, replacing MobileNet\cite{2020MobileNets} with SCPDarknet \cite{Bochkovskiy2020} to serve as the backbone. Li \textit{et al.} \cite{Li2024} employed the heatmap as output of deep neural network to extract the skyline. Ahmad \textit{et al.} \cite{Ahmad2017} took a comparison between four sky segment approaches including Edge-less horizontal detection (DCSI) \cite{Ahmad2015Edgeless}, Automatic Labeling Environment (ALE) \cite{Saurer2016}, Fully Convolutional Network (FCN) \cite{Long_2015_CVPR} and SegNet \cite{Badrinarayanan2017}. In this comparing work, the validation test indicated the performance of FCN is best for sky segmentation of mountainous imagery. The deep learning methods are mainly designed based on deep neural networks, which usually contain a large amount of parameters. Their robustness is better than those traditional methods. Simultaneously, the more labeled data is indispensable for the training of deep neural networks.

\section{METHODOLOGY}
\begin{figure*} 
    \centering
    \includegraphics[width=18cm]{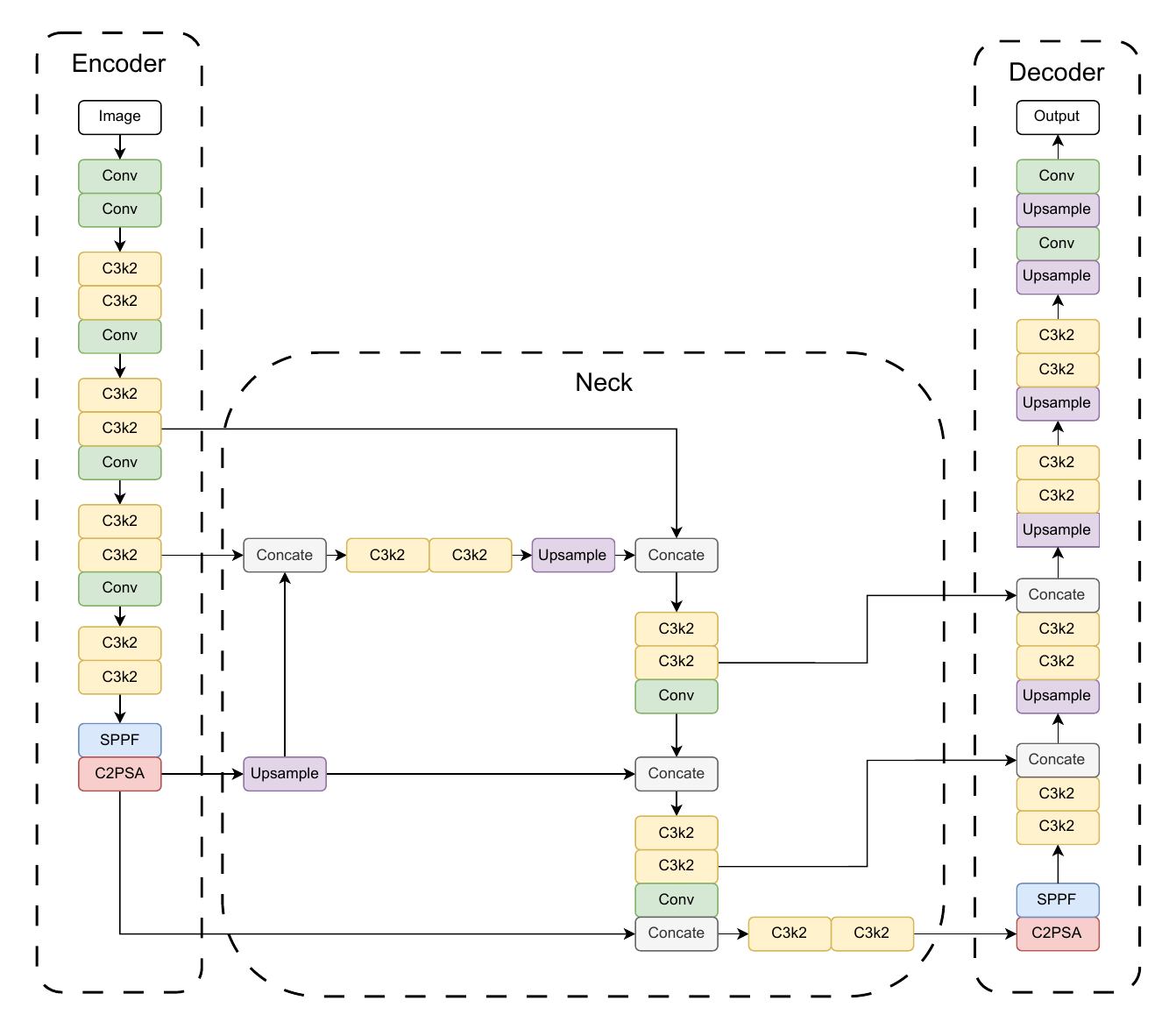}
    \caption{overall of YUNet}
    \label{fig:my_label}
\end{figure*}

\par In this research, we designed YUNet, a deep learning-based skyline detection algorithm. The YUNet is designed based on YOLOv11 architecture for better sky segmentation and skyline detection performance. And then based on the segmentation results, the skyline can be extracted by some simple edge extraction algorithm. Sky segmentation is the key innovation that can directly influence skyline extraction performance. Therefore, the YUnet can perform the sky semantic segmentation task, predicting each pixel in the given image whether it belongs to the sky region or not.

The YOLOv11 architecture is a famous deep learning framework for object detection, but it cannot perform the image segmentation task. The YOLOv11 architecture consists of backbone, neck, and head submodule. The backbone and neck are used for feature extraction and multi-scale feature fusion, respectively. The head is designed for predicting the classes and bounding boxes of objects. Benefiting from the neck between the backbone and head, the YOLOv11 architecture can make a fusion of multi-scale features, the YOLOv11 can collect more large-range contextual and multi-scale information. 
\par
As mentioned in the Introduction, the UNet architecture are limited to obtain large-range contextual information to handle complex sky segmentation tasks. In order to capture more multi-scale and large-range contextual features, we designed the YUNet built from YOLOv11 architecture.
The YUNet is an UNet-like architecture. but it has a neck submodule for fusion of the large-range contextual information, so that the shortage of original UNet can be solved.
\par
The YUNet is designed by three levels: Firstly, we take the backbone of YOLOv11 as the encoder for feature extraction. Secondly, the head network in YOLOv11 architecture is designed for predicting the classes and bounding boxes of objects, which is not necessary for image segmentation tasks. We remove it from the original YOLOv11 architecture. Thirdly, in order to generate the final output of the sky segmentation task, we designed a decoder submodule to replace the head network. As two symmetric paths, the encoder and decoder have the same number of layers, which is designed for the preservation of spatial information and balanced paths for image reconstruction. 
\par Thus, the novel architecture of the encoder-neck-decoder is designed as shown in Fig. 1. The encoder extracts and downsamples the representations from the given image by a cascade of convolutional layers. As the downsampling of feature maps, the spatial information will be lessened, while the semantic information will be enriched, gradually. The neck makes the fusion of low-level spatial and high-level semantic features from the encoder. The decoder uses the fused features from neck to reconstruct the segmentation map by a series of convolution and upsample layers.

\section{Expriments}
We adopt two evaluation methods for this research, the sky segment and skyline detection evaluation.
The sky segment evaluation is performed directly on the mask of prediction and ground truth, focusing on the evaluation of each mask pixel prediction. The sky segment evaluation is performed on the Skyfinder dataset, which is split into the train and validation dataset. The skyline detection evaluation is performed on the boundary between sky and not-sky regions, focusing on the evaluation of the skyline shape prediction. The performance of the sky segment evaluation contains Dice Score, IoU, and MCR \cite{s21216996}. While the performance of the skyline detection evaluation contains $\mu$, $\sigma$, \textit{Min}, and \textit{Max} \cite{Ahmad2021}. For the skyline detection evaluation, the model is trained on the GeoPose3k dataset and validated on the CH1 dataset.
\subsection{Datasets}
\textbf{Skyfinder dataset}. This dataset is built from 90,000 long-term timelapse images from 53 outdoor webcams over a variety of lighting and weather conditions. Its images contain a wide range of sizes and aspect ratios, including 640×489, 857×665,960×600, 1280×720, and 1280×960. The binary mask is provided at each location. In this research, the Skyfinder dataset is divided into two parts. The first containing 77535 images and masks is used for training, and the last containing 13914 images and masks is used for testing.
\par
\textbf{CH1 dataset}. This dataset is proposed for geo-localization by Saurer\cite{Saurer2016}. It contains 165 images and masks. all of whose size is processed as 1024×679. In this research, the edge between the sky and non-sky region is extracted by the Canny algorithm for skyline detection evaluation. All images of the CH1 dataset are used for testing the model trained from the Skyfinder dataset.
\par

\subsection{Parameter set}
We adopt the YOLOv11 as our baseline.  SGD optimizer with momentum of 0.937, learning rate of 0.01, and weight decay of 0.0005 was initialized in training process. We trained all models in 50 epochs. We used single GTX4090Ti for training with batch size of 4 and prediction with single.

\subsection{Comparision with the State-of-the-art}
\textbf{Results on Skyfinder.} For evaluation of the sky segmentation of YUnet, we report the full experimental results on the test part of the Skyfinder dataset, as shown in Table \uppercase\expandafter{\romannumeral1}, where the YUNet-n, YUNet-s, YUNet-m, YUNet-l, and YUNet-x is developed from the VanillaNet-N, VanillaNet-S, VanillaNet-M, VanillaNet-L, and VanillaNet-X respectively. The comparison in accuracy, precision, recall, dice score, and IoU shows the YUNet we proposed has an art-of-state performance, which is better than other famous image segmentation methods. The visualization of performance comparison in complex environments is shown in Fig.2. The visualization shows that in complex environments, such as bad lightness or weather, YUnet is more stable than other methods outlined in Table \uppercase\expandafter{\romannumeral1}. We observed the segmentation becomes more difficult near the boundary between the sky and the non-sky region. The YUNet still performs better than others, especially near the boundary between the sky and the non-sky region.  

\begin{table*}[!t]

\caption{Performance comparison on Skyfinder dataset}
\normalsize
    \centering
    \setlength{\tabcolsep}{3.5mm}{
    \begin{tabular}{lccccc}
    \hline
        Method           &Accuracy(\%) &Precision(\%) &Recall(\%)  &Dice-Score(\%)     &IoU  \\
    \hline
    U-Net\cite{Ronneberger2015}  &98.285 &98.464 &98.094 & 98.033 &0.9614\\   
    NI-UNet\cite{s21216996}  &99.232 &99.211 &99.221 &99.104 &0.9822\\
    SegNet\cite{Badrinarayanan2017} &99.415 &98.851 &99.104 &98.914 &0.9799 \\
    FCNs\cite{Long_2015_CVPR} &99.181 &98.316 &98.905 &98.528 &0.9731 \\
    YOLOv11-seg\cite{YOLO_2023} &99.389 &99.417 &98.435 &98.906 &0.9787\\
    Adaptive NN\cite{2020Sky}  &-- &94.6 &96.5 &-- &0.952 \\
    \hline
    YUNet-n(ours) &99.195 &98.308 &98.834 &98.513 &0.9719 \\
    YUNet-s(ours) &99.467 &99.169 &98.969 &99.042 &0.9816 \\
    YUNet-m(ours) &99.479 &99.382 &98.809 &99.078 &0.9821 \\
    YUNet-l(ours) &99.406 &99.186 &99.034 &99.036 &0.9824 \\
    YUNet-x(ours) &99.565 &99.465 &99.101 &99.253 &0.9858 \\
    \hline
    \end{tabular}}
    \label{tab:my_label}
\end{table*}
\textbf{Results on CH1 dataset.} For evaluation of the skyline detection of YUnet, we report the experiments of YUnet-x trained from the GeoPose3K dataset and tested on CH1 datasets. In skyline detection evaluation, we employ pixel-wise absolution distance (PAD) as our metrics. The PAD\cite{Ayadi2016} values were obtained by calculating the skyline vector using the Sobel edge extraction algorithm on the output of our model. The PAD can be calculated using the following equation. Here, we use E to denote PAD.
\begin{equation}
    E = \frac{1}{N}\sum_{i=1}^{n}\lvert p_{i}-p_{i}^{*} \rvert
\end{equation}
where \begin{math}p_{i}\end{math} and \begin{math}p_{i}^{*}\end{math} denote skyline pixels at the i-th column of the predicted result and ground truth. \begin{math}N\end{math} is the number of the image column. Table \uppercase\expandafter{\romannumeral2} shows the index statistical results for different methods on the CH1 dataset. Shallow Learning\cite{Ahmad2021} is a kind of simple filter-learning method. DCNN-DSI\cite{2016A}, DeepLabv3\cite{Chen2017RethinkingAC}, FCNs\cite{Long_2015_CVPR}, PSPNet\cite{Zhao_2017_CVPR} are DNN-based segmentation methods. MSSDN\cite{DBLP} is a DNN-based skyline predict method. The listed experiment results illustrate that DNN-based methods have a higher accuracy compared with the case of using a simple classifier for skyline detection. In comparison of listed DNN-based segmentation methods, YUNet we proposed has a state-of-the-art performance.
\par We observed that MSSDN is slightly better than YUNet because MSSDN\cite{DBLP} is not a segmentation model. It predicts skyline probability graphs directly. YUNet is a segmentation model that predicts the skyline with post-processing, such as Sobel, Canny, or other edge extraction algorithms. The post-precessing can introduce some deviation into predicted results. While the direct predicting models are affected by category imbalance in their training process. These models are also disturbed by variety and noise, including poor lightness and harsh weather.       

\begin{table}[!t]
\caption{Performance comparison on CH1 dataset}
\normalsize
\centering
    \setlength{\tabcolsep}{1.5mm}{
    \begin{tabular}{|c|c|c|c|c|} 
        \hline
        \multirow{2}{*}{Method}& \multicolumn{4}{c|}{CH1 dataset} \\ 
        \cline{2-5}   
        &$\mu$ &$\sigma$ &Min &Max \\
        \hline
        Shallow Learning\cite{Ahmad2021} &38.28 &59.52 &0.80 &349.06 \\
        \hline
        DCNN-DSI\cite{2016A} &4.51 &14.06 &0.45 &14.0\\
        \hline
        DeepLabv3\cite{Chen2017RethinkingAC} &2.06 &0.70 &1.22 &8.10\\
        \hline
        FCN8s\cite{Long_2015_CVPR} &1.75 &2.00 &0.64 &20.52\\
        \hline
        FCN16s\cite{Long_2015_CVPR} &1.96 &1.21 &0.84 &13.18 \\
        \hline
        FCN32s\cite{Long_2015_CVPR} &3.17 &1.59 &1.73 &22.06\\
        \hline
        PSPNet\cite{Zhao_2017_CVPR} &1.35 &0.89 &0.45 &8.90 \\
        \hline
        MSSDN\cite{DBLP} &1.25 &0.65 &0.43 &3.92 \\
        \hline
        
        YUNet &1.36 &1.05 &0.44 &9.24 \\
        \hline
    \end{tabular}}
 \label{tab:my_label}
\end{table}

\begin{figure*} 
    \centering
    \includegraphics[width=18cm]{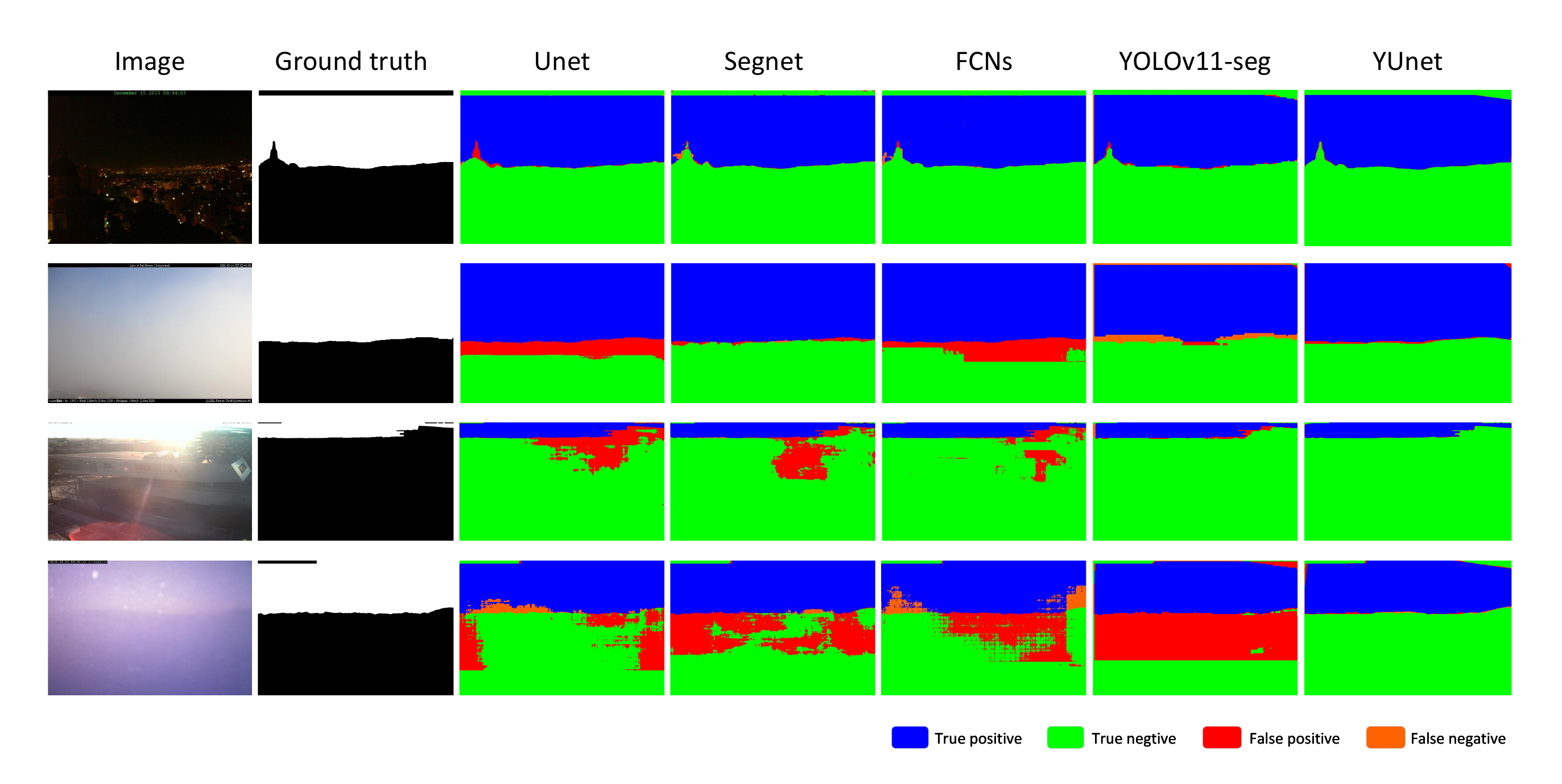}
    \caption{visualization of comparison}
    \label{fig:my_label}
\end{figure*}

\section{Conclusion}
The YUNet is proposed for sky region segmentation or skyline detection in this paper. The YUNet based on YOLOv11 has a novel encoder-neck-decoder architecture. The main advantage of YUNet is high detection accuracy and robust in the complex environment including bad lightness and harsh weather. The evaluation experiments on our YUNet and other mainstream methods are performed on Skyfinder and CH1 dataset. The experiments show that YUNet achieved a state-of-the-art results when compared with other methods. In further, we will research more advanced methods under touger conditions.



 
%

\bibliographystyle{IEEEtran}
\bibliography{IEEEabrv, ref}



\section{Biography Section}
 
\vspace{-22pt}

\begin{IEEEbiography}[{\includegraphics[width=1in,height=1.25in,clip,keepaspectratio]{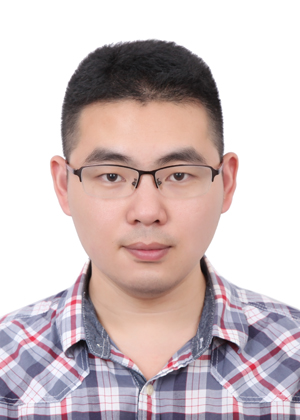}}]{Gang Yang}
received his B.S. and M.S. degrees from Beijing Jiaotong University in 2014 and 2019, respectively. He is working for Beijing Institute of Survey and Mapping and Beijing Key Laboratory of Urban Spatial Information Engineering, in Beijing, China. His research interests include machine learning and remote sensing image processing. (yanggang3510@gmail.com).
\end{IEEEbiography}

\vspace{11pt}

\begin{IEEEbiography}[{\includegraphics[width=1in,height=1.25in,clip,keepaspectratio]{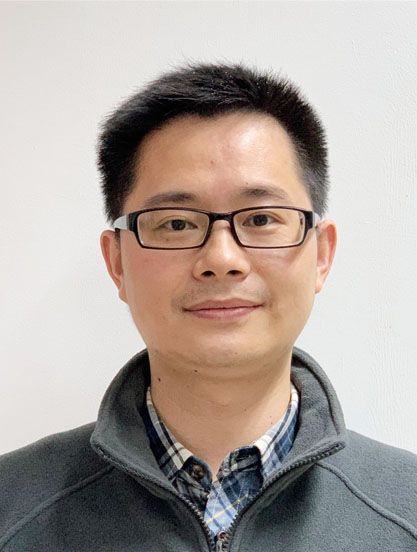}}]{Miao Wang}
received his B.S. and M.S. degrees from Wuhan University, China, in 2008 and 2010, respectively. Since 2010, he has been working for Beijing Institute of Surveying and Mapping as a senior engineer. His research interests are in urban spatial analysis and geographic information science. (wangmiao@bism.cn).
\end{IEEEbiography}

\vspace{11pt}

\begin{IEEEbiography}[{\includegraphics[width=1in,height=1.25in,clip,keepaspectratio]{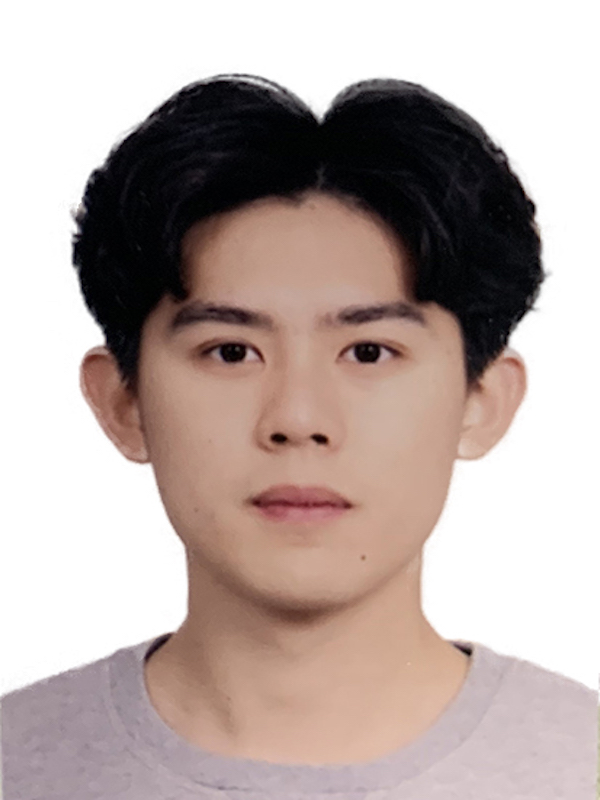}}]{Quan Zhou}
received his B.S. degree from Capital Normal University, China, in 2015 and his M.S. degree from the University of Twente, Netherlands in 2018. Since 2018  he has been an engineer at Beijing Institute of Surveying and Mapping.  His research interests are in smart city, remote sensing, and system architecture design.
\end{IEEEbiography}

\vspace{11pt}

\begin{IEEEbiography}[{\includegraphics[width=1in,height=1.25in,clip,keepaspectratio]{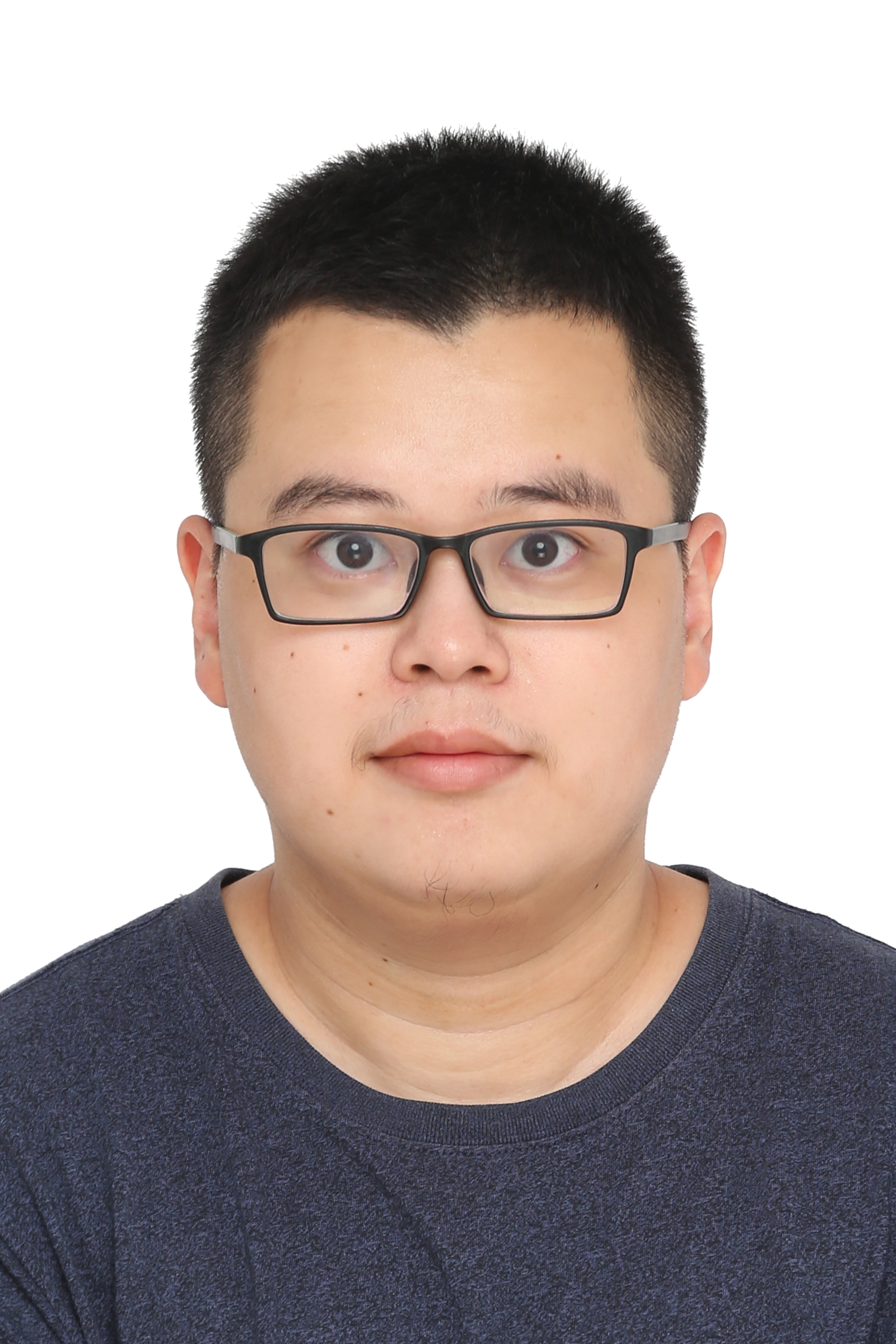}}]{Jiangchuan Li}
received his B.S. and M.S. degrees from Beijing University of Civil Engineering and Architecture, China, in 2019 and 2022, respectively. Since 2022, he has been an assistant engineer at Beijing Institute of Surveying and Mapping. His research interests are in GIS and remote sensing.
\end{IEEEbiography}

\vfill

\end{document}